\documentclass[letterpaper, 10 pt, journal, twoside]{ieeetran}
\usepackage[pdftex]{graphicx}
\usepackage{tabularx}
\usepackage{multirow}

\usepackage{amsmath}
\usepackage{amssymb}
\usepackage{amsfonts}

\usepackage{balance}

\ifCLASSOPTIONcompsoc
 \usepackage[caption=false,font=normalsize,labelfont=sf,textfont=sf]{subfig}
\else
 \usepackage[caption=false,font=footnotesize]{subfig}
\fi

\usepackage{url}
\usepackage{hyperref}			% Managing cross-referencing

\begin{document}
\title{Cooperative Aerial-Ground Multi-Robot System for Automated Construction Tasks}

\author{Marko Krizmancic, Barbara Arbanas, Tamara Petrovic, Frano Petric and Stjepan Bogdan% <-this % stops a space
\thanks{Manuscript received: September 10, 2019; Revised: December 2, 2019; Accepted: December 29, 2019}
\thanks{This paper was recommended for publication by Editor Nak Y. Chong upon evaluation of the Associate Editor and Reviewers’ comments.}
\thanks{Authors are with University of Zagreb, Faculty of Electrical Engineering and Computing, Unska 3, 10000 Zagreb, Croatia{\tt\small (marko.krizmancic, barbara.arbanas, tamara.petrovic, frano.petric, stjepan.bogdan) at fer.hr}}% <-this % stops a space
\thanks{Digital Object Identifier (DOI): \href{https://doi.org/10.1109/LRA.2020.2965855}{10.1109/LRA.2020.2965855}}}


\markboth{IEEE ROBOTICS AND AUTOMATION LETTERS. PREPRINT VERSION. ACCEPTED DECEMBER 2019}%
{Krizmancic \MakeLowercase{\textit{et al.}}: Cooperative Aerial-Ground Multi-Robot System for Automated Construction Tasks}

\maketitle

\begin{abstract}
In this paper, we study a cooperative aerial-ground robotic team and its application to the task of automated construction. We propose a solution for planning and coordinating the mission of constructing a wall with a predefined structure for a heterogeneous system consisting of one mobile robot and up to three unmanned aerial vehicles. The wall consists of bricks of various weights and sizes, some of which need to be transported using multiple robots simultaneously. To that end, we use hierarchical task representation to specify interrelationships between mission subtasks and employ effective scheduling and coordination mechanism, inspired by Generalized Partial Global Planning. We evaluate the performance of the method under different optimization criteria and validate the solution in the realistic Gazebo simulation environment.
\end{abstract}


% \fi
\IEEEpeerreviewmaketitle

% Input the content.
\section{Introduction}
\label{sec:introduction}
% Intro
\IEEEPARstart{I}{n} the field of robotics, the focus of the research is shifting increasingly from the design and control of single robots towards collaborative methods for multi-robot systems. Especially interesting are heterogeneous systems since the diverse capabilities of robots in such systems foster better performance, broader space coverage, improved energy utilization, and better knowledge through data fusion. This is particularly true for aerial-ground systems due to their distinct skillsets. For instance, unmanned aerial vehicles (UAVs) may benefit from ground vehicle's extended battery life while unmanned ground vehicles (UGVs) could take advantage of far-ranging mobility of aerial robots \cite{Petrovic2015, Arbanas2018}.

The majority of the works related to UAV-UGV teams consider cooperation in terms of: collaborative sensing, data fusion and information sharing between UAV and UGV, such as collaborative mapping of the environment \cite{Qin2019, Potena2019}, reconnaissance and surveillance \cite{Langerwisch2013, Maini2018}, multi-robot localization \cite{DeSilva2014}, collaborative object manipulation \cite{Staub2017}, target tracking \cite{Yu2015}, as well as search and rescue \cite{Marconi2013}.

% State of the art
Controlling a heterogeneous team of robots requires precise high-level task planning and fast and robust coordination mechanisms. Given the nature of the task and environment at hand, a decentralized approach to a multi-robot coordination problem is bound to give the best performance concerning robustness and resilience to potential robot fall-outs. Some of the most prominent solutions to the problem mentioned above are an auction and market-based approaches \cite{Gerkey2002, Liu2013a}. These mostly fit into the task allocation category, where a set of simple tasks is given, and robots use bidding mechanisms to distribute tasks amongst themselves. Application of bidding mechanisms for more complex missions, in an uncertain environment with heterogeneous agents, is given in \cite{ElGibreen2019}, however partial ordering between tasks and tight coupling, which are the basis of our cooperative missions, are not well supported.

Probabilistic multi-robot coordination approaches based on usage of decentralized partially observable Markov decision processes have been studied as well \cite{Amato2015, Floriano2019}. The advantage of this approach is its inherent suitability to uncertain environments; however, the scalability problem is making them unsuitable for real-world applications that include several robots and complex tasks. Although some online solutions have been reported (\cite{Wu2011}), the problems this approach manages to cope with are still relatively simple from the reasoning perspective.

Many of recent state-of-the-art approaches rely on off-the-shelf automated reasoners based on, for example, Linear Temporal Logic (LTL) \cite{Gombolay2018, Schillinger2017}. Even though these approaches show significant contributions to the theoretical synthesis of correct-by-design controllers, they often suffer from intensive computational problems, as well as the inability to quantify planner objectives. In \cite{DeCastro2018}, the authors devised a method for the automatic synthesis of controllers for a team of robots working in a dynamic environment. The advantage of this approach is the ability to cope with the dynamic constraints of the robot and guarantees collision avoidance in 2D and 3D workspaces. The method's performance is evaluated on a simple task of garbage collection using humanoid robots. However, LTL-based approaches remain limited by the expressivity of the logic, and it is challenging to specify complex task relations.

% Our approach
The approach presented herein entails dynamic high-level task planning (decomposition and allocation), which appropriates T\AE MS (Task Analysis, Environment Modeling, and Simulation) language \cite{taems1999, Lesser2004} for task representation. The advantage of T\AE MS task representation is the ability to easily specify complex task interrelations (both tight and loose) and to quantify the potential solutions. It, however, requires a planner and coordinator able to exploit all the benefits of such an expressive task model. We devised a system able to do just that \cite{Arbanas2018}, based on general guidelines on the design of the communication and coordination protocol of the Generalized Partial Global Planning (GPGP) coordination framework \cite{decker1995}. In our previous work, we applied the described method to different multi-robot missions, such as disaster response \cite{Petrovic2015} and autonomous parcel delivery \cite{Arbanas2018, Arbanas2016}.

Contributions of this work are following. Firstly, we extend the abilities of the framework regarding task allocation by implementing a market-based protocol, for the class of problems ST-MR-TA-XD (single-task robots, multi-robot tasks, time-extended assignment, cross-schedule dependencies) \cite{korsahTaxonomy}. Consequently, we can cope with complex scheduling problems in real-time, without drawbacks on the optimality of obtained solutions. Further, the protocol also involves temporal and precedence constraints, which is not the usual case in the state-of-the-art literature. We apply the method to the problem of automated construction, which perfectly highlights the ability of the planner to cope with combinatorial optimization problem of heterogeneous multi-robot systems in complex missions. The main advantage of our approach is the ability to clearly and effectively quantify goals of multi-criteria optimization function, as demonstrated in the results section. Moreover, to the best of our knowledge, none of the state-of-the-art approaches provide a complete solution to the described decentralized problem, but rather focus only on one aspect of the problem, mainly (centralized) task allocation \cite{Gombolay2018, Nunes2017}. Lastly, since the planner employs a mission-agnostic approach, this solution can be applied to any domain with minimal alterations, as long as task representation adheres to the required hierarchical structure.

% The paper is structured as follows
The paper is structured as follows. In Section \ref{sec:problem_description}, we describe the problem in detail. Section \ref{sec:method} entails the specifics of the used method, while Section \ref{sec:simulation_results} shows simulation results using the described system. Finally, in Section \ref{sec:conclusion}, we give a brief conclusion and comments about future work.

\section{Problem description}
\label{sec:problem_description}

% MBZIRC
The \textit{Mohamed Bin Zayed International Robotics Challenge} (MBZIRC)\footnote{\href{http://mbzirc.com}{http://mbzirc.com}} aims to encourage innovation and highest quality research in emerging topics by providing a demanding set of robotics challenges that require robots working more autonomously in dynamic, unstructured environments while collaborating and interacting with each other. The technological challenges addressed in the MBZIRC 2020 Challenges include fast autonomous navigation in semi-unstructured, complex, dynamic environments, with reduced visibility (e.g. smoke) and minimal prior knowledge, robust perception and tracking dynamic objects in 3D, sensing and avoiding obstacles, GPS denied navigation in indoor-outdoor environments, physical interactions, complex mobile manipulations, and air-surface collaboration \cite{mbzirc}. In this paper, we tackle the problem of automated construction where a team of aerial and ground robots needs to collaborate to autonomously locate, pick, transport and assemble various types of brick-like objects to build predefined structures. Particularly, we are interested in the decision making and coordination process which optimizes wall construction task concerning given multi-objective criteria.

% MBZIRC challenge
The challenge specification is given as follows. Red, green, blue, and orange bricks ranging from 0.3m to 1.8m in length are separated by color and assembled in randomly located piles before the start of the challenge. Blue bricks may be collected only by the UAVs, while orange bricks need to be carried by two or more UAVs at the same time due to their size and weight. Either UAVs or the UGV may assemble red and green bricks, but those constructed by the UAVs will gain higher marks. The form of the wall structure is the only input to the system. Teams have 30 minutes to complete the challenge and score is determined by the percentage and speed of completion. Exact parameters of the challenge and brick properties are further specified in \cite{mbzirc}.

% Robot collaboration
This challenge has been specifically designed to emphasize the need for collaboration between different types of agents. UAVs are agile, have a virtually unlimited reach and are superior in number, which means they can construct more in less time. However, due to their limited battery life, it is imperative to also utilize the UGV for low-scoring bricks and those closer to the ground.

% Goal
The focus of this paper is the implementation of an algorithm for decentralized high-level mission planning and coordination which will supervise the task of construction. All available agents should be utilized to their full potential and carry out assigned tasks in parallel to maximize the challenge score and complete the task within the specified time.

\section{Method}
\label{sec:method}

% TAEMS
\subsection{Model description}
We decompose the wall building mission using T\AE MS hierarchical task structure (\cite{Petrovic2015}, \cite{Lesser2004}) in a way that the immediate subtasks of root task represent transportation and assembly of individual bricks. 
The T\AE MS tree is defined for three types of agents, "UGV", "UAV", and "UAVx2", where the last label denotes an agent, comprised of two UAVs that are capable of collaborating in the transport of the large, orange brick.  Leafs of the tree represent \textit{action nodes} that correspond to real, actionable agent behaviours. Agent behaviours used in presented applications are $GP(b_i)$ - go to the pile and locate the brick $b_i$, $PU(b_i)$ - pick up the brick from the pile, $GW(b_i)$ - go to the wall, and $PD(b_i)$ - place the brick in its specified location within the wall, where $i$ signifies brick identifier. Other nodes are \textit{task nodes} that combine action nodes into logical units. A task node that combines all actions necessary to transport a single brick $b_i$ is denoted with $TB(b_i)$.

Four types of relations between nodes define how quality, duration, and cost of children nodes affect the parent node -- \textit{q\_sum\_all} that corresponds to logical AND, \textit{q\_max} that represents the XOR operator, \textit{q\_seq\_sum\_all} that is AND with strictly sequential execution, and \textit{q\_sum} which signifies that a subset of subtasks renders parent task as finished. In T\AE MS language, these relations are called \textit{quality accumulation functions} or $QAF$ for short.
\textit{Quality} of each task represents the number of points received for successfully assembling the individual brick. \textit{Duration} of a task is estimated using the length of the desired path and average moving speed of the agent. Additionally, for actions $PU(b_i)$ and $PD(b_i)$ a fixed estimated duration of grabbing and releasing behaviours is added. Finally, we determine the \textit{cost} by multiplying duration with the per-time cost based on the complexity of the task, type of the agent performing it, and the size of the associated brick.

If a task cannot be completed due to kinematic constraints of the robot (e.g. desired placement of the brick is unreachable by the UGV's robotic arm), quality of such task is set to zero, while duration and cost are set to large positive number in order to exclude it from scheduling.

% IR
Relations between other action and task nodes are modelled using interrelationship T\AE MS elements. 
\textit{Enables} relations demand that one task finishes before another is started and are used to specify the order of bricks within the wall structure.
The action of picking up one brick ($PU(b_i)$) enables the action of going to the pile to pick up another brick ($GP(b_j)$). 

This introduces an offset in the transportation of neighboring bricks and helps to avoid situations where multiple agents are placing bricks at the same time in close proximity. The offset is introduced at the beginning of tasks because it is more efficient for the UAVs to wait on the ground than in the air while carrying a brick.

\textit{Disables} relations are used to further prioritize the order of the brick placements in order to ensure the feasibility of the schedules.

% Simple redundancy
\subsection{Resolving simple redundancy}
One of the most essential steps in the GPGP coordination procedure is a resolution of simple redundancy between tasks. Redundancy is a situation where one or more agents in the set denoted with $A_{sr}$ may execute a single task. Resolution procedure starts by selecting one of the agents as a referee. The selected agent then collects task assessments from all other agents, chooses the best one and communicates the result with others. Agents that are not selected as the best discard the task from their schedules.

The first part of the criteria function for selecting the best agent to execute a redundant task $j$ is defined as follows:
\begin{align}
    r_{Q,i}(j) &= \frac{Q_i(j) - Q_{min}(j)}{Q_{max}(j) - Q_{min}(j)}, \\
    r_{D,i}(j) &= \frac{D_{max}(j) - D_i(j)}{D_{max}(j) - D_{min}(j)}, \\
    r_{C,i}(j) &= \frac{C_{max}(j) - C_i(j)}{C_{max}(j) - C_{min}(j)}, \\
    R_i(j) &= \alpha \cdot r_{Q,i}(j) + \beta \cdot r_{D,i}(j) + \gamma \cdot r_{C,i}(j), \label{eq:Ri}
\end{align}
where $Q_i(j)$ is the quality of task $j$ assessed by the agent $i \in A_{sr}$, $Q_{min}(j) = min_{\, i \in A_{sr}}(Q_i(j))$ and $Q_{max}(j) = max_{\, i \in A_{sr}}(Q_i(j))$. Analogous definitions are given for $D_i(j)$, $D_{min}(j)$ and $D_{max}(j)$ which represent the duration assessment, and $C_i(j)$, $C_{min}(j)$ and $C_{max}(j)$ for the cost. Parameters $\alpha$, $\beta$ and $\gamma$ are user-defined positive real constants, $\alpha + \beta + \gamma = 1$, that give different importance to various criteria.

The problem arises when redundant tasks or agents are very similar. In this case, task assessments are almost identical, but usually, one agent has a marginally better outcome and is selected to execute all redundant tasks. Other agents are therefore not utilized, and tasks cannot execute in a parallel manner. In order to better allocate the redundant tasks, we utilize additional market-based criteria function, as described in the following paragraph.

Based on initial duration assessments and interrelationships between tasks, a simple market-based task allocation function creates an allocation scheme in a way which aims to minimize the total mission duration. The result of the function is a list $S$ of pairs in the form of $(task,\ agent \ assigned \ to \ the\ task)$. Such allocation is a preferred solution for the redundancy ($r_2$) but task assessments of each agent ($r_1$) are still used to cover the cases where some agent has significantly better execution outcome. The total rating of the agent $i$ for the given task $j$ is:
\begin{align}
    r_{1,i}(j) &= \frac{R_i(j) - R_{min}(j)}{R_{max}(j) - R_{min}(j)}, \\
    r_{2,i}(j) &= \begin{cases}
    1, & \text{if $(j,i) \in S$,} \\
    0, & \text{otherwise,}
    \end{cases} \\
    R_{total,i}(j) &= \delta \cdot r_{1,i}(j) + (1-\delta) \cdot r_{2,i}(j),
\end{align}
where $R_i(j)$ is a rating of the task $j$ based on quality, duration, and cost assessment of the agent $i \in A_{sr}$, as defined in (\ref{eq:Ri}), $R_{min}(j) = min_{\, i \in A_{sr}}(R_i(j))$, $R_{max}(j) = max_{\, i \in A_{sr}}(R_i(j))$, and $\delta$ is a user-defined positive constant, $\delta \in [0.5,1)$, that gives different importance to the two parts of the criteria function. Note that for $\delta < 0.5$, rating $r_1$ would not affect the total rating because $r_{1,i}(j) \le r_{2,i}(j),\ \forall i \in A_{sr}$ and $\sum_{i \in A_{sr}}r_{2,i}(j) = 1$. In other words, for each task $j$, there would be only one agent for which $r_{2,i} = 1$ and that agent would have better total rating than all others, regardless of the value of $r_1$.

% Complex redundancy
\subsection{Resolving complex redundancy}
Complex redundancy is an extension to the T\AE MS model with the addition of a so-called local $QAF$. It represents the situation where a task has multiple subtasks that need to be executed simultaneously and therefore, it is necessary to have a one-on-one mapping between subtasks and agents. Such a situation occurs in the case of the task of joint transportation of orange brick. Two agents must schedule their part of the task at the same time and execute it together. Globally, for the task to be executed, both subtasks must be performed, while locally to each agent, only one of those tasks is to be scheduled. Therefore, tasks are modeled as complexly redundant by defining the $QAF$ of their parent task as \textit{q\_sum} or \textit{q\_sum\_all} and $QAF_{local}$ as \textit{q\_max}. We use the term redundancy since the effect is similar to one of the simply redundant tasks, but its resolution needs to be managed differently. Instead of multiple agents competing for a single task, as is the case during the resolution of simple redundancy, $m$ agents compete for $n$ tasks, $m \ge n$, such that each agent schedules only one subtask and that all subtasks are assigned.

The problem of assigning only one agent to each task in a way that optimizes total quality, duration and cost is modelled as a generalized assignment problem \cite{Kundakcioglu2009}:
\begin{align}
    &\mathrm{max} & &\sum_{i \in A_{cr}} \sum_{j \in T_{cr}} R_i(j) \cdot x_{ij}, \\
    &\mathrm{subject\ to} & &\sum_{i \in A_{cr}} x_{ij} = 1, \ \forall j \in T_{cr}, \label{eq:taskagent}\\
    & & &\sum_{j \in T_{cr}} x_{ij} = 1, \ \forall i \in A_{cr}, \label{eq:agenttask}
\end{align}
where $A_{cr}$ is a set of agents that are able to schedule a complexly redundant task, $T_{cr}$ is a set of subtasks of the complexly redundant task, $R_i(j)$ is the rating of the subtask $j$ based on assessment of the agent $i$ defined in equation (\ref{eq:Ri}) and $x_{ij} \in \{0,1\}$ is a binary decision variable for the assignment of agent $i$ to the subtask $j$. Equations (\ref{eq:taskagent}) and (\ref{eq:agenttask}) ensure that each subtask is assigned to only one agent and that each agent performs only one subtask respectively. Described optimization problem is solved using the custom implementation of the branch and bound algorithm \cite{Little1963}.

% Resources
\subsection{Resources}
Each set of action nodes connects to its parent task with a \textit{q\_seq\_sum\_all} function, meaning that they must carry out in a strictly specified order. However, this does not prevent an agent from scheduling and executing other tasks in the meantime.

\begin{figure}[!t]
    \centering
    \includegraphics[width=0.45\textwidth]{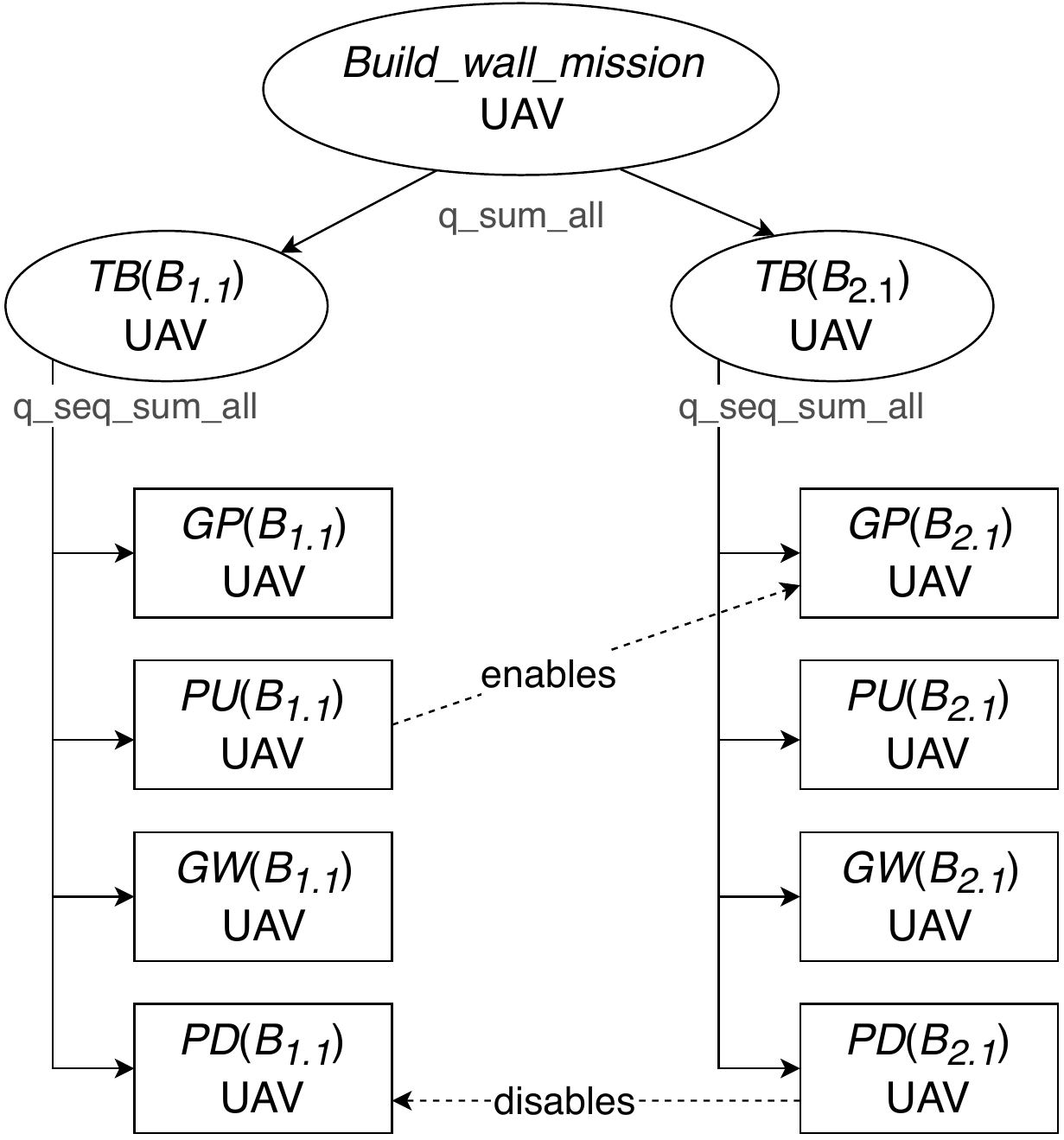}
    \caption{Task structure of the example mission}
    \label{fig:ex1}
\end{figure}

Consider a system with only one UAV and two blue bricks that need to be stacked on top of each other. T\AE MS tree structure for such mission is shown in Figure \ref{fig:ex1}. The bottom brick is labeled as $B_{1.1}$, while the top one is labeled $B_{2.1}$. Expected execution schedule for specified mission is as follows:
\begin{align*}
&GP(B_{1.1}),\ PU(B_{1.1}),\ GW(B_{1.1}),\ PD(B_{1.1}),\\
&GP(B_{2.1}),\ PU(B_{2.1}),\ GW(B_{2.1}),\ PD(B_{2.1}).
\end{align*}

However, successful execution of action $PU(B_{1.1})$ enables action $GP(B_{2.1})$ and final execution schedule is in fact: 
\begin{align*}
&GP(B_{1.1}),\ PU(B_{1.1}),\ GP(B_{2.1}),\ PU(B_{2.1}), \\ 
&GW(B_{1.1}),\ GW(B_{2.1}),\ PD(B_{1.1}),\ PD(B_{2.1}). 
\end{align*}

\begin{figure}[!t]
    \centering
    \includegraphics[width=0.45\textwidth]{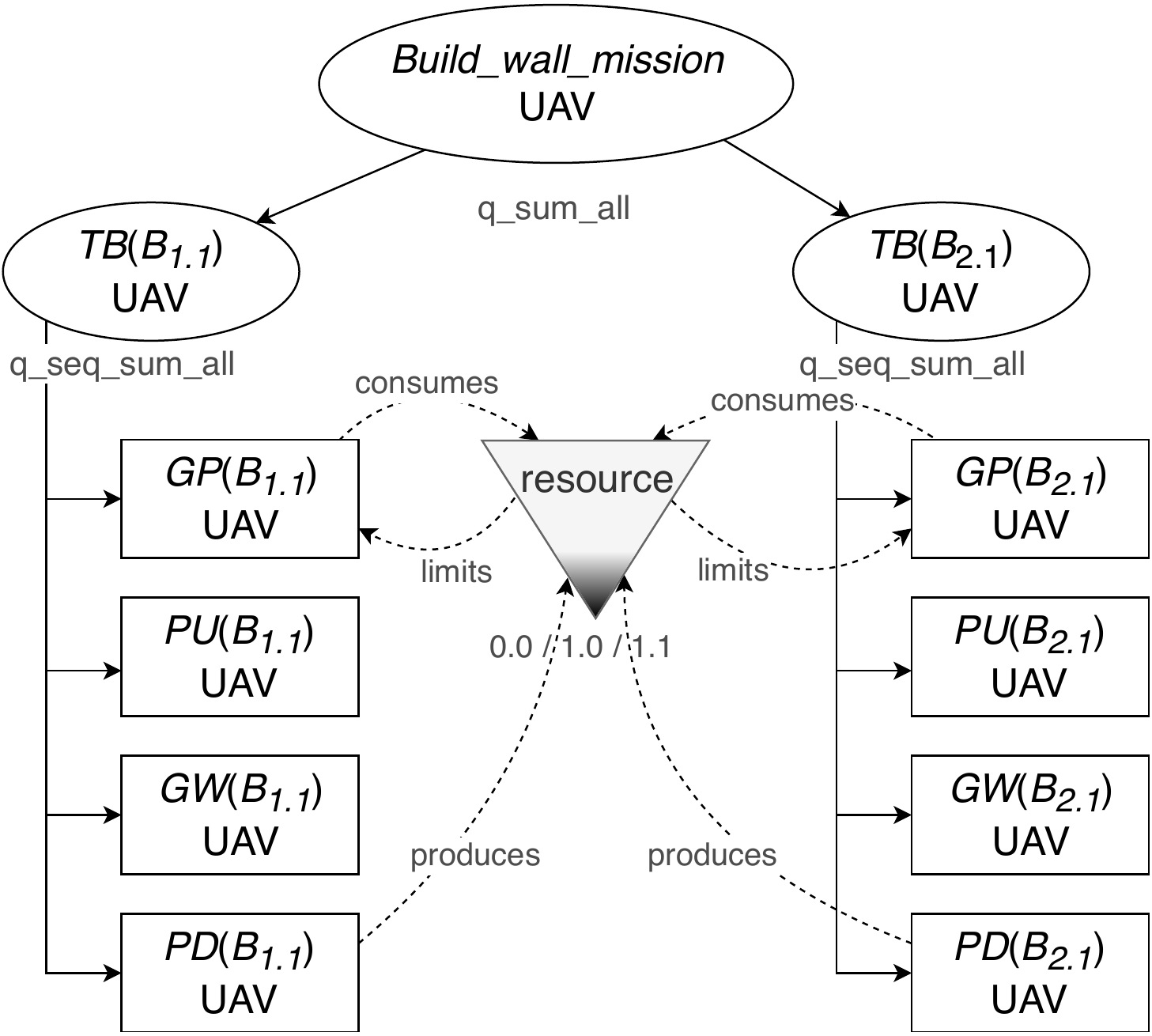}
    \caption{Task structure of the example mission extended with resources. \textit{Enables} relations are omitted for clarity.}
    \label{fig:resource}
\end{figure}

The scheduling algorithm is intentionally unaware of the agents' specific abilities, so it suits various applications. The final schedule is the best solution for the given problem because it minimizes the total mission duration, even though UAVs cannot carry multiple bricks at the same time.
In order to model such limitation, T\AE MS virtual resources are used in the extended model of the mission. Each agent entails a resource, which represents a slot for carrying a brick. The initial state of the resource is $1$, while its lower and upper limits are $0$ and $1.1$ respectively. Every time an agent executes $GP(b_i)$ type of action, one unit of resource is consumed, making it insufficient.
Insufficient resources automatically disable all other $GP(b_j),\ (i \neq j)$ actions for that agent until their state is restored by executing the $PD(b_i)$ action.
The improved model of T\AE MS tree structure is shown in Figure \ref{fig:resource}.

\section{Simulation results}
\label{sec:simulation_results}

\subsection{Testbed description}
\begin{figure}[!t]
    \centering
    \includegraphics[width=0.48\textwidth]{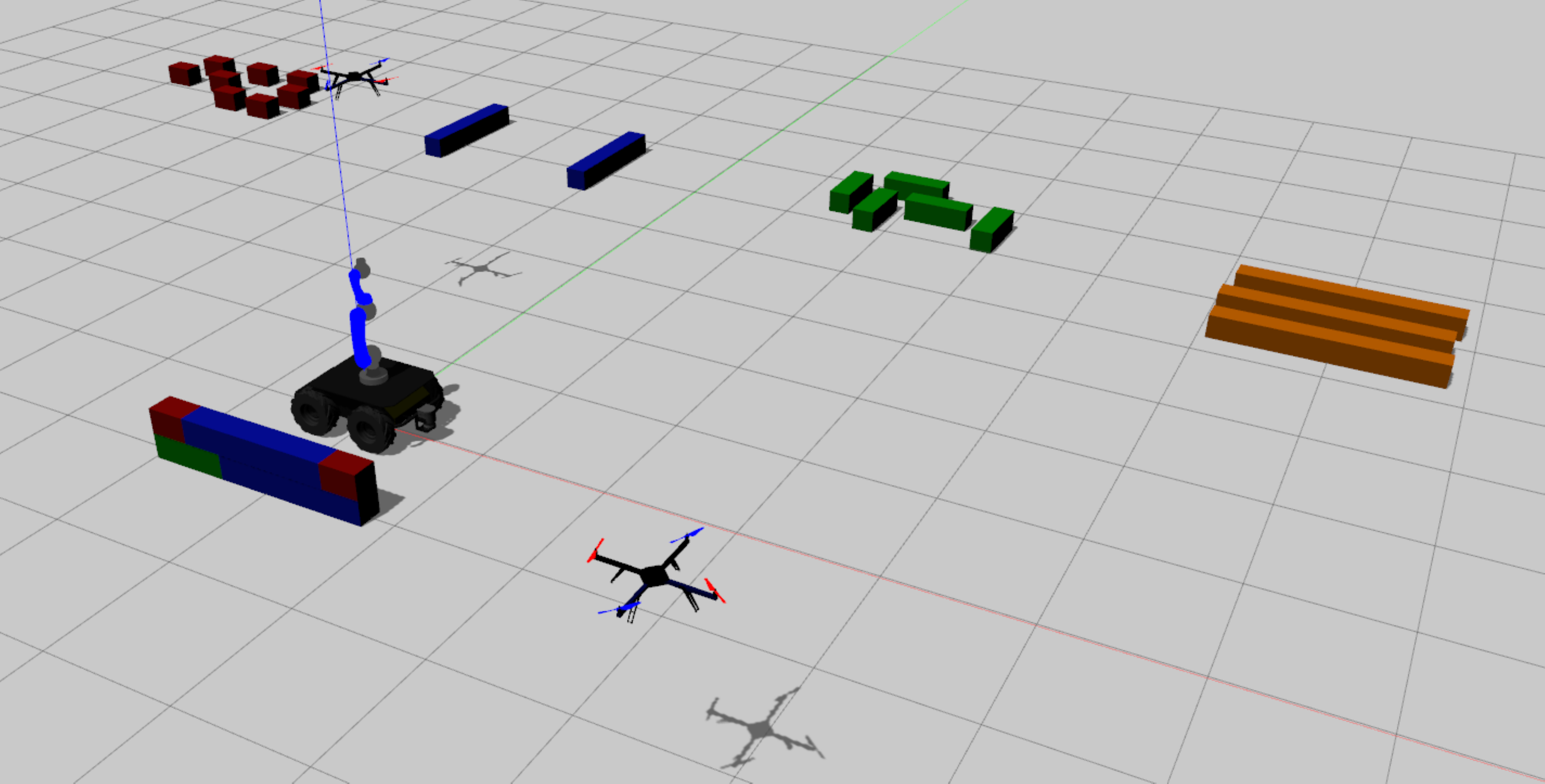}
    \caption{Simulation environment with robots}
    \label{fig:gazebo}
\end{figure}

We devised the environment for the second challenge of MBZIRC 2020 in the \textit{Gazebo} simulator (Figure \ref{fig:gazebo}). Using its \textit{Robot Operating System} (ROS) interface, we can realistically simulate planning, coordination, and execution aspects of the proposed solution for wall building mission.

We simulate physically accurate models of multicopters for UAV agents and a Gazebo model of Husky by Clearpath Robotics with mounted Schunk Powerball LWA 4P robotic arm for the UGV agent. Both models have standard sensors like inertial measurement unit and a generic pose sensor which provide true position and orientation of the robots. UAVs use a simple position PID controller for movement, while UGV uses a navigation algorithm provided by \textit{move\_base} ROS package. 
Mid-air collisions are prevented by commanding a different predefined flying height to each UAV.

Since the focus of this work is on planning and coordination, robots do not have end effectors for interaction with the bricks onboard. In the simulation, bricks are teleported using Gazebo's \textit{spawn} and \textit{delete} services.

Locations of all four brick piles and wall origin are known before the start of the mission. In the real world, challengers should acquire this information by scouting the arena with UAVs at the start of the trial. 

\subsection{Simulation results}
\begin{figure}[!t]
    \centering
    \includegraphics[width=0.35\textwidth]{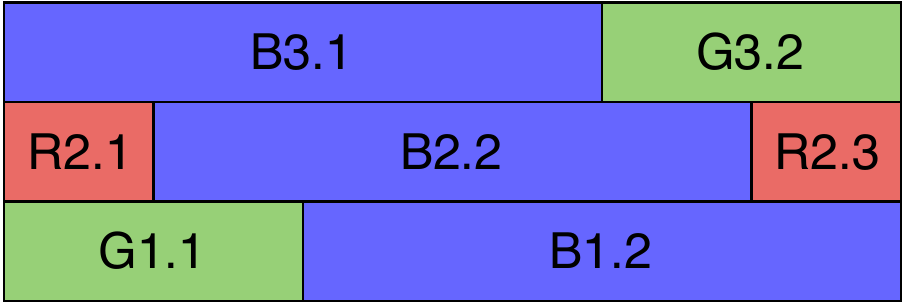}
    \caption{Specified wall structure for the first mission}
    \label{fig:mission1}
\end{figure}

In order to test the system described in this paper, we employ the team consisting of two UAVs and a UGV on a task specified as in Figure \ref{fig:mission1}. We conducted the simulated mission with a series of different criteria setups, three of which we analyze here. The costs of execution per time unit are kept constant for UAVs and UGV in all of the scenarios.

\begin{figure}[!t]
    \centering
    \includegraphics[width=0.42\textwidth]{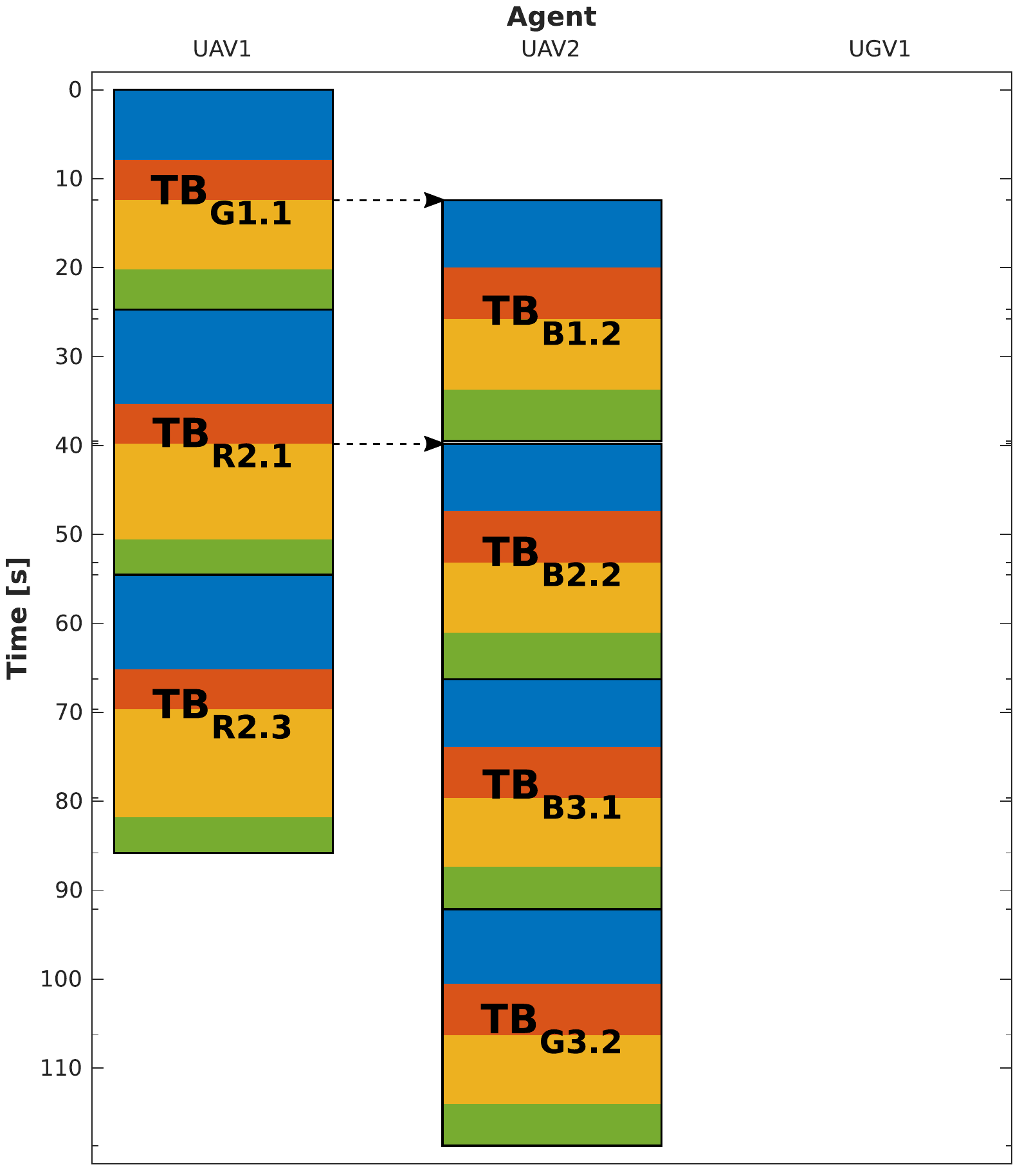}
    \caption{Execution schedule of the first mission for the criteria defined as $\alpha=0.5, \beta=0.35, \gamma=0.15$}
    \label{fig:mission_q}
\end{figure}

In the first scenario, we define the total quality of the mission as the most weighty component of the planning criteria while other components play a less significant role, $\alpha=0.5, \beta=0.35, \gamma=0.15$. As expected, only UAVs participate in the construction because bricks assembled in that way carry more points. Our task allocation method assigns tasks to both UAVs alternately in order to achieve parallel execution and shorten the total mission duration. Time diagram of tasks executed by each agent is shown in Figure \ref{fig:mission_q}. Colored segments split each task $TB(b_i)$ into its actions, where blue corresponds to action $GP$, red to $PU$, yellow to $GW$, and green to $PD$. Dashed arrows represent \textit{enables} relations which affect the final schedule.

\begin{figure}[!t]
    \centering
    \includegraphics[width=0.42\textwidth]{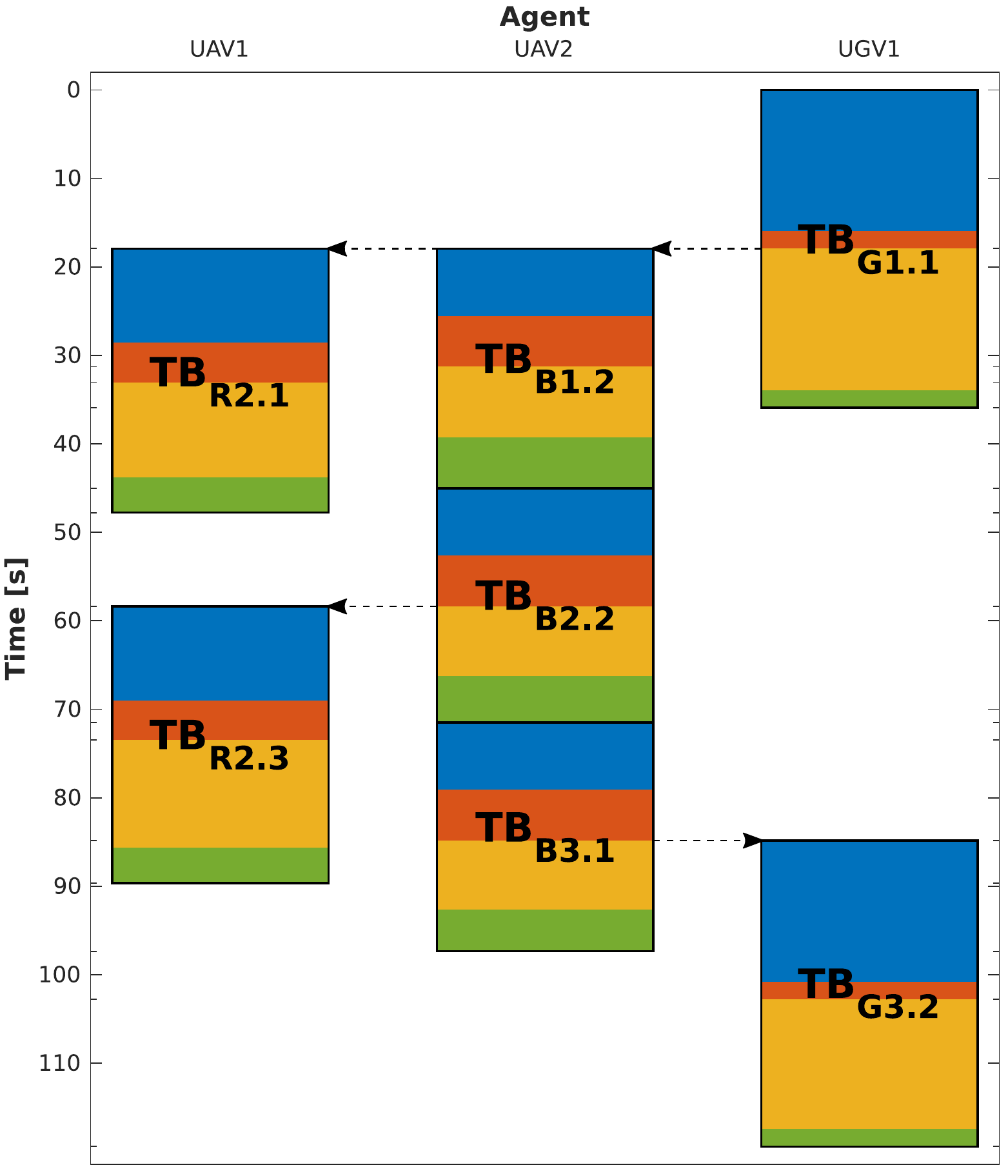}
    \caption{Execution schedule of the first mission for the criteria defined as $\alpha=0.35, \beta=0.15, \gamma=0.50$}
    \label{fig:mission_cqd}
\end{figure}

In the second scenario, the most weight is given to the cost component of the criteria $\alpha=0.35, \beta=0.15, \gamma=0.50$. Because the UGV has a lot smaller cost per time unit than the UAVs, tasks related to green bricks are assigned to the UGV. We have defined that red bricks provide 100\% more points when assembled by UAVs, as opposed to 40\% for the green bricks. Since the quality of the mission is still somewhat important, a better outcome is achieved if the UAVs assemble red bricks. The execution schedule for this scenario is shown in Figure \ref{fig:mission_cqd}.

\begin{figure}[!t]
    \centering
    \includegraphics[width=0.42\textwidth]{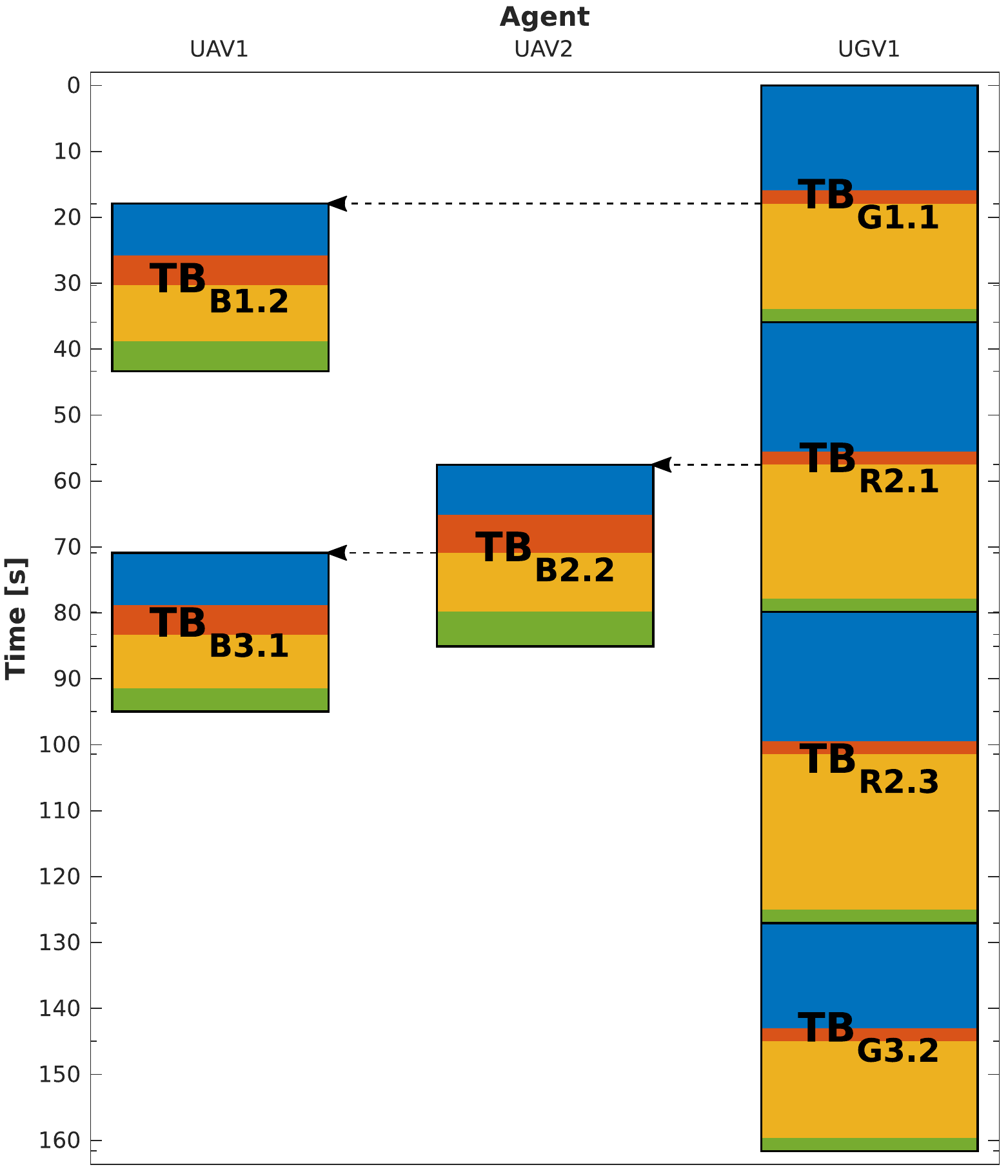}
    \caption{Execution schedule of the first mission for the criteria defined as $\alpha=0, \beta=0, \gamma=1$}
    \label{fig:mission_c}
\end{figure}

Figure \ref{fig:mission_c} shows the execution schedule of the third scenario which is fully focused on minimizing the total cost of the mission. The criteria is defined as $\alpha=0, \beta=0, \gamma=1$. In this case, tasks related to red bricks are also assigned to the UGV since their increased score potential does not affect the mission outcome. Blue bricks are still assembled by the UAVs as is defined in the challenge description.

\begin{figure}[!t]
    \centering
    \includegraphics[width=0.4\textwidth]{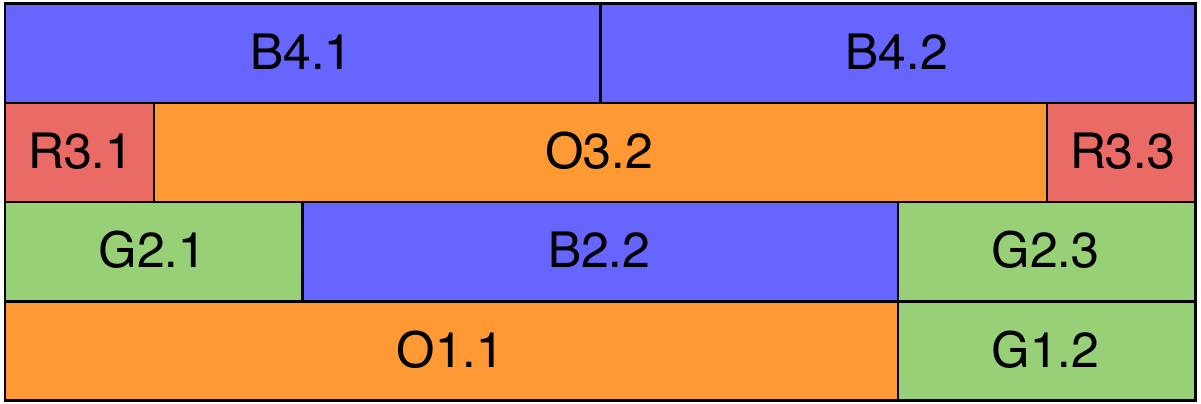}
    \caption{Specified wall structure for the second mission}
    \label{fig:mission2}
\end{figure}

Lastly, we add a third UAV and employ the team of robots on a more complex mission specified as in Figure \ref{fig:mission2}.
In order to reduce the mission duration, but still include the UGV in the execution, we use the same criteria as in the second scenario of the first mission, $\alpha=0.35, \beta=0.15, \gamma=0.5$.

\begin{figure}[!t]
    \centering
    \includegraphics[width=0.45\textwidth]{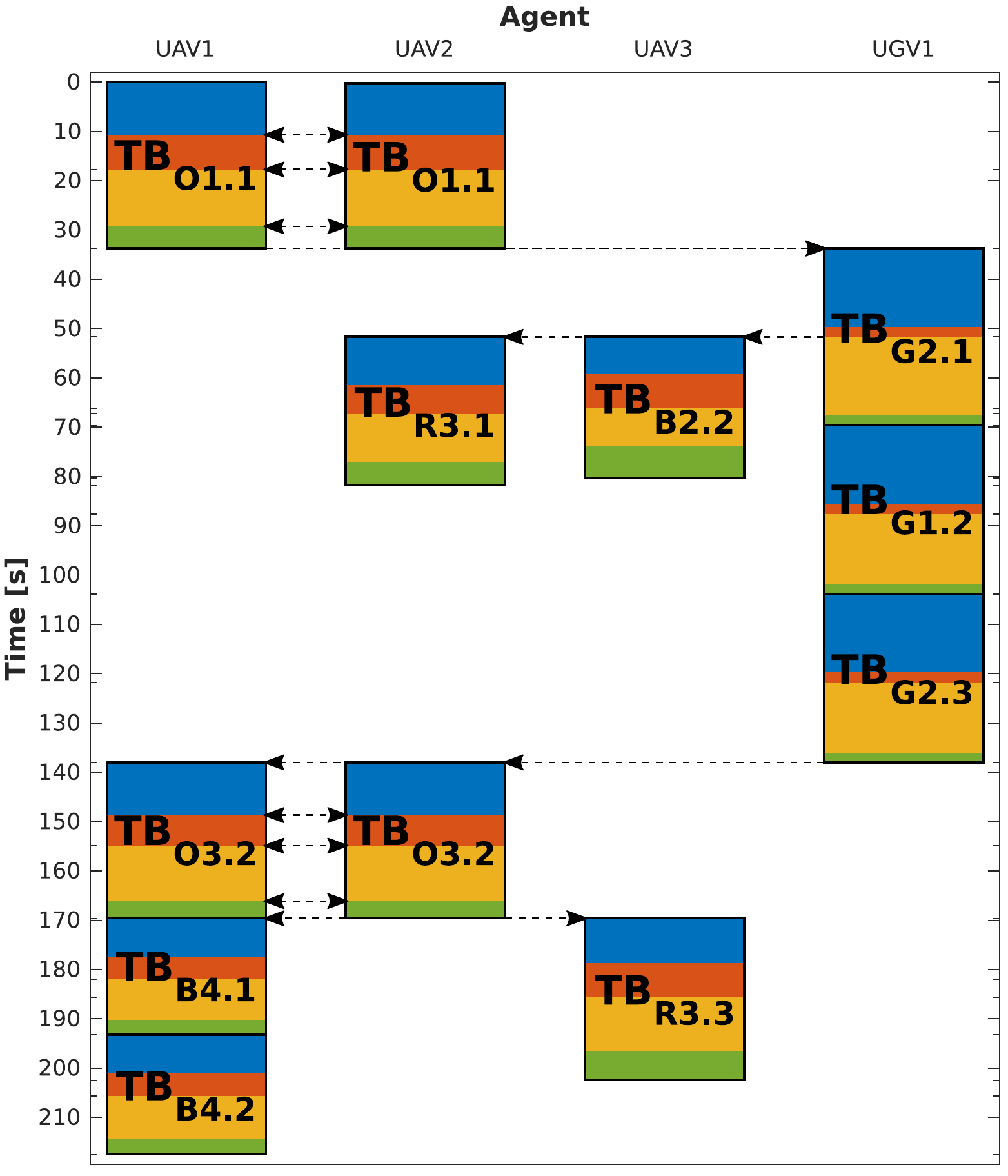}
    \caption{Execution schedule of the second mission}
    \label{fig:mission2_result}
\end{figure}

Generated execution schedule for the last mission is shown in Figure \ref{fig:mission2_result}. As in the previous scenarios, tasks are evenly distributed between agents in order to shorten the total mission duration and reduce overall cost while maximizing the score. Results confirm that the proposed method is capable of producing optimized high-level plans for different agent and criteria setups, and that it meets current challenge requirements.

We conducted 20 additional simulations for problems of similar dimensionality and compared the performance with solutions obtained by the Gurobi Optimizer \cite{gurobi}. Gurobi Optimizer uses exact mathematical methods for solving mixed integer linear programming (MILP) problems and for the specified mission set always returns an optimal solution. We also compare generated schedules with the state-of-the-art iterated auction-based approach with a single central agent acting as an auctioneer \cite{Nunes2017}. We tested each of the 20 mission scenarios on a set of five different criteria specifications and recorded the average value and the standard deviation of the objective function in respect to the optimal solution. The results of the experiments are shown in the Table \ref{tab:res}.

Results show that the proposed method on average performs within 12\% of the optimum for the majority of tested criteria specifications, with the exception of the criteria fully oriented on minimizing the mission makespan. During the resolution of simple redundancy, tasks are always assigned to faster agents (e.g. UAVs), meaning that agents with significantly worse duration assessments (e.g. UGV) do not participate in the mission execution, thus prolonging the total duration.

Auction-based approach performs slightly better in terms of the achieved optimality gap, however, our approach is fully decentralized and has the ability to handle scheduling and simultaneous execution of multi-agent tasks, e.g. transportation of orange bricks.
Another major advantage of the proposed method is that it is mission-agnostic and only a hierarchical task structure is needed, contrary to the Gurobi Optimizer which requires an exact mathematical formulation. Furthermore, mathematical methods used by Gurobi often fail to provide a valid solution when applied to more complex problems.

\begin{table}[ht!]
\caption{Optimality gaps of the proposed solution and auction based approach compared to Gurobi Optimizer. Lower values are better.}
\label{tab:res}
\centering
\begin{tabular}{c|c|c|c}
\hline
\multirow{2}{*}{Criteria}                                & \multirow{2}{*}{} & Proposed & \multirow{2}{*}{Auction} \\
                                                         &                   & solution &                          \\ \hline
\multirow{2}{*}{$\alpha=0.5,\ \beta=0.35,\ \gamma=0.15$} & $\mu$             & 11.15 \% & 8.18 \%                  \\
                                                         & $\sigma$          & 6.26 \%  & 6.00 \%                  \\ \hline
\multirow{2}{*}{$\alpha=0.35,\ \beta=0.15,\ \gamma=0.5$} & $\mu$             & 6.43 \%  & 8.10 \%                  \\
                                                         & $\sigma$          & 10.96 \% & 3.82 \%                  \\ \hline
\multirow{2}{*}{$\alpha=1,\ \beta=0,\ \gamma=0$}         & $\mu$             & 0 \%     & 0 \%                     \\
                                                         & $\sigma$          & 0 \%     & 0 \%                     \\ \hline
\multirow{2}{*}{$\alpha=0,\ \beta=1,\ \gamma=0$}         & $\mu$             & 42.38 \% & 2.18 \%                  \\
                                                         & $\sigma$          & 11.54 \% & 4.187 \%                 \\ \hline
\multirow{2}{*}{$\alpha=0,\ \beta=0,\ \gamma=1$}         & $\mu$             & 2.61 \%  & 0 \%                     \\
                                                         & $\sigma$          & 1.62 \%  & 0 \%                     \\ \hline
\end{tabular}
\end{table}

An example of collaboration between two UAVs during the assembly of the large, orange brick is also examined. Due to the complexity of this task, other tasks are temporarily suspended during its execution in order to avoid collisions and various disturbances. \textit{Enables} relations are therefore placed between $PD$ and $GP$ actions of each orange and its neighbouring bricks.

\begin{figure}[!t]
    \centering
    \includegraphics[width=0.48\textwidth]{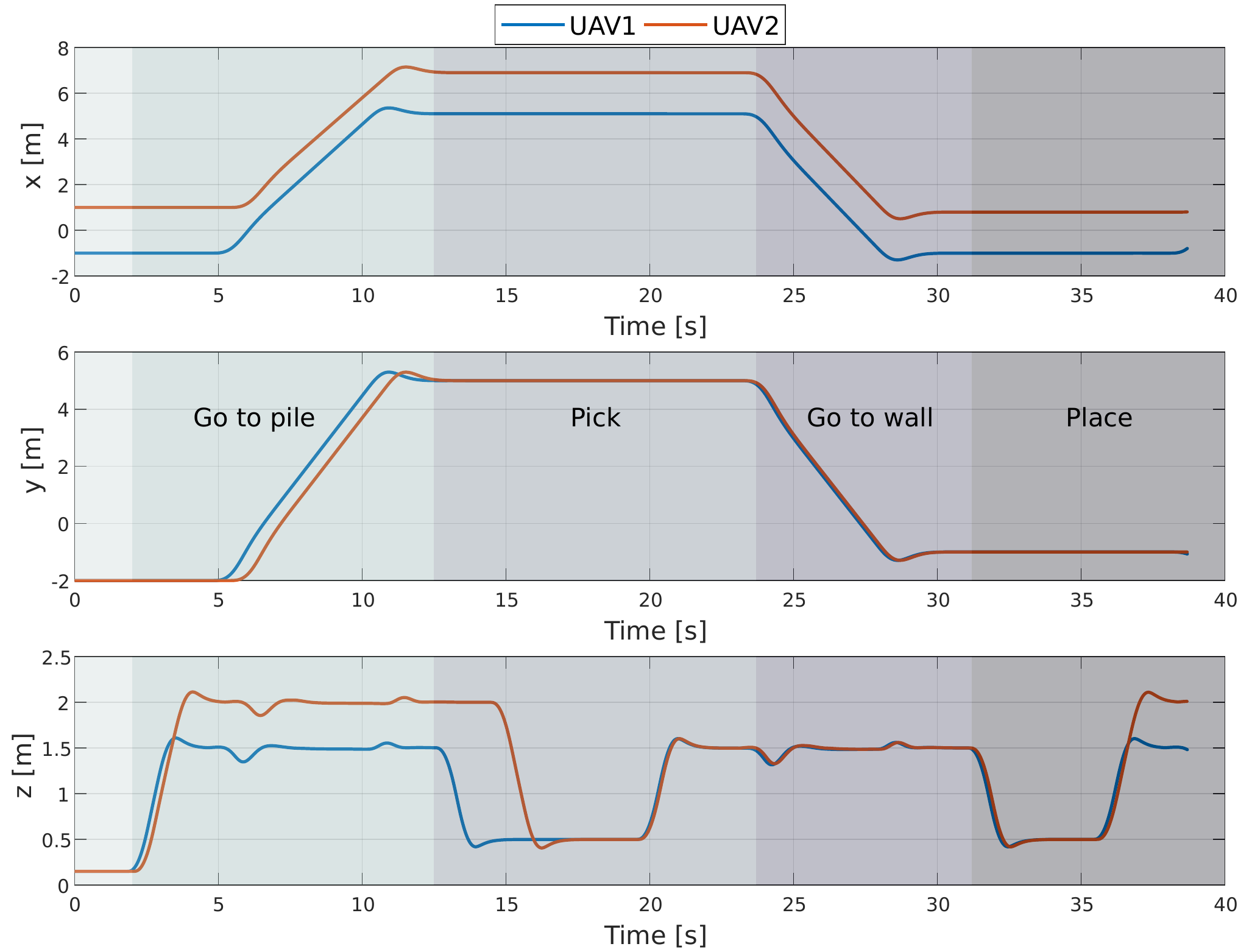}
    \caption{Trajectories of UAVs during collaborative transportation of the orange brick}
    \label{fig:collaboration}
\end{figure}

Figure \ref{fig:collaboration} shows the trajectories of both UAVs during the execution of the described task. The figure is plotted concerning time, clearly marking agents' specific actions (takeoff, going to position, picking and placing the brick). Synchronization of the UAVs is achieved by modelling their tasks as complexly redundant and interconnecting action nodes with \textit{enables} relations. During execution, participating agents form a direct communication link to share their status and further synchronize all steps of the currently executing action.

Video of the described simulations can be found at \cite{video}. We also applied the  method to the mission of constructing a wall with a larger number of bricks to highlight the scalability of the proposed solution. The figure of the resulting schedule is too large to be conveniently included in the paper itself, but it is available at \cite{wiki}.
\section{Conclusion and future work}
\label{sec:conclusion}

In this paper, we have proposed and tested a decentralized task planning and coordination framework for a cooperative aerial-ground robotic team on a case study of the task of automated construction. The team needs to collaborate to autonomously locate, pick, transport and assemble various types of brick-like objects to build predefined structures. The approach assumes that the location of bricks and hierarchical decomposition of system mission are known a priori.
We simulate physically accurate models of multicopters for UAV agents and a Gazebo model of Husky by Clearpath Robotics with mounted Schunk Powerball LWA 4P robotic arm for the UGV agent. Results show that our method can successfully generate schedules for cooperative missions of tightly coupled tasks under various setups of a multi-criteria objective.

As future work, we aim to implement and test the developed solution on a real aerial-ground robotic system used in the MBZIRC 2020 competition. Further, we plan to improve the method in terms of robustness and resilience to failed attempts of highly risky task execution with the inclusion of on-line re-planning capability. Finally, we plan to investigate the possible integration of dynamic task assignment algorithms into the already developed framework to make it even more applicable in real life situations.

\appendices

\section*{Acknowledgment}
This work has been supported by European Commission Horizon 2020 Programme through project under G. A. number 810321, named Twinning coordination action for spreading excellence in Aerial Robotics - AeRoTwin and the Mohammed Bin Zayed International Robotics Challenge. The work of doctoral student Barbara Arbanas has been supported in part by the “Young researchers’ career development project—training of doctoral students” of the Croatian Science Foundation funded by the European Union from the European Social Fund.

\ifCLASSOPTIONcaptionsoff
  \newpage
\fi



%\bibliographystyle{IEEEtran}
\bibliographystyle{IEEEtran}
\balance
\bibliography{bibliography.bib}

\end{document}